\documentclass[10pt,twocolumn,letterpaper]{article}

\usepackage[pagenumbers]{wacv} %

\usepackage{graphicx}
\usepackage{amsmath}
\usepackage{amssymb}
\usepackage{booktabs}

\usepackage[pagebackref,breaklinks,colorlinks]{hyperref}

\usepackage{epsfig}
\usepackage{graphicx}
\usepackage{amsmath}
\usepackage{amssymb}
\usepackage{pifont}
\usepackage{booktabs}
\usepackage{multirow}
\usepackage[position=top]{subcaption}
\usepackage{makecell}
\usepackage{colortbl}
\usepackage{dsfont}
\usepackage{comment}
\usepackage{xcolor}

\usepackage{nicefrac}

\usepackage{algorithm}
\usepackage[noend]{algpseudocode}

\usepackage{bm}
\usepackage{pifont}

\usepackage[capitalize]{cleveref}
\crefname{section}{Sec.}{Secs.}
\Crefname{section}{Section}{Sections}
\Crefname{table}{Table}{Tables}
\crefname{table}{Tab.}{Tabs.}

\newcommand{\tablestyle}[2]{\setlength{\tabcolsep}{#1}\renewcommand{\arraystretch}{#2}\centering\footnotesize}
\newcommand{\tablestylesmaller}[2]{\setlength{\tabcolsep}{#1}\renewcommand{\arraystretch}{#2}\centering\scriptsize}
\renewcommand{\paragraph}[1]{\vspace{1.25mm}\noindent\textbf{#1}}

\newcommand{\uline}[1]{\underline{#1}}
\definecolor{baselinecolor}{gray}{.9}

\def\wacvPaperID{1681} %
\def\confName{WACV}
\def\confYear{2025}

\newcommand{\affiliationskip}{\hspace{2.5mm}}

\begin{document}

\title{Generalized Category Discovery under the Long-Tailed Distribution}

\author{Bingchen Zhao\textsuperscript{\textcolor{red}{\ding{169}}} \quad\quad Kai Han\textsuperscript{\ding{171}}\textsuperscript{\textdagger}\\
{\tt\small zhaobc.gm@gmail.com \quad\quad kaihanx@hku.hk} \\
\vspace{.5em}
\textsuperscript{\textcolor{red}{\ding{169}}}  University of Edinburgh \affiliationskip \textsuperscript{\ding{171}} The University of Hong Kong
}

\maketitle
\renewcommand{\thefootnote}{\fnsymbol{footnote}}
\footnotetext[2]{Corresponding author.}

\begin{abstract}
This paper addresses the problem of Generalized Category Discovery (GCD) under a long-tailed distribution, which involves discovering novel categories in an unlabelled dataset using knowledge from a set of labelled categories. Existing works assume a uniform distribution for both datasets, but real-world data often exhibits a long-tailed distribution, where a few categories contain most examples, while others have only a few. While the long-tailed distribution is well-studied in supervised and semi-supervised settings, it remains unexplored in the GCD context. We identify two challenges in this setting - balancing classifier learning and estimating category numbers - and propose a framework based on confident sample selection and density-based clustering to tackle them. Our experiments on both long-tailed and conventional GCD datasets demonstrate the effectiveness of our method.
\end{abstract}

\section{Introduction}
\label{sec:intro}

Over the recent years, computer vision has shown significant advancements in tasks such as image recognition~\cite{resnet}. Despite these progressions, artificial systems still face challenges in recognizing and categorizing visual information accurately in dynamic and complex environments. The visual information present in the real world is far more diverse and intricate than the benchmark datasets. To tackle this issue, researchers have directed their attention towards learning techniques that require minimal human intervention, such as the semi-supervised learning approach~\cite{oliver2018realistic}.
However, one limitation of most semi-supervised methods is the common assumption that the unlabelled dataset contains a set of categories with a small labelled dataset.
This assumption is unrealistic as it is not possible to label all categories in the real world at once, not to mention the categories in the unlabelled dataset may grow over time.

\begin{figure}[t]
\centering
    \includegraphics[width=1.0\columnwidth]{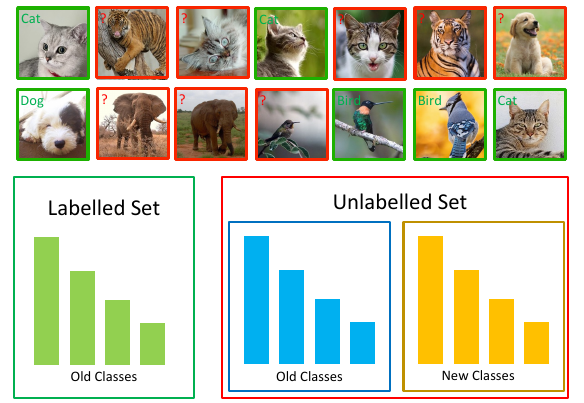}
\vspace{-5pt}
\caption{
Generalized Category Discovery (GCD) under a long-tailed distribution: Given labelled images from seen categories and unlabelled images from seen and unseen categories, the objective is to automatically assign labels to the unlabelled images, where the distribution in the labelled and unlabelled data is long-tailed.
}
\label{fig:teaser}
\vspace{-20pt}
\end{figure}

Therefore, efforts have increasingly focused on the task of generalized category discovery, where the model's primary objective is not only to recognize the known categories in the labeled dataset but also to identify the novel categories within the unlabeled dataset. This paper takes this research one step further by exploring generalized category discovery under a more realistic setting, where the labeled and unlabeled datasets have a long-tailed distribution, meaning that a few categories have a substantial number of examples (head classes), while the other categories contain only a few instances (tail classes) (see~\cref{fig:teaser}). The primary challenge in this context lies in the potential bias towards the head classes, making it difficult to identify and accurately recognize the tail classes. Although this long-tailed setting has been highly explored under the fully and semi-supervised settings, to our knowledge, no prior research has been conducted to address the long-tailed distribution in generalized category discovery.

This paper proposes a new approach to tackle generalized category discovery under the long-tailed distribution. We address two challenges associated with this task through the following proposed methods:

First, we introduce a novel method that adaptively selects confident samples from the unlabelled dataset using prediction confidence and density. These identified samples are utilized to form training mini-batches to balance the distribution of the model's training and to reduce bias. Additionally, the prediction distribution of the model to be close to a prior distribution of those confident samples, further helps in mitigating the bias.

Second, under the long-tailed distribution, previous approaches \cite{Han2019learning,vaze22generalized} for estimating the novel class number may be imprecise as they rely on the $k$-means algorithm, which has shown to be inaccurate when used for long-tailed distributions\cite{liang2012k,wu2009adapting}. Therefore, this work proposes a unique density-peak-based class number estimation method that is insensitive to imbalances in the data distribution.

In summary, the contributions of our paper are as follows:

\begin{itemize}
\item We extend the generalized category discovery to the long-tailed distribution setting and showcase the unique challenges that long-tailed distribution presents, such as learning an unbiased representation and accurately estimating the number of classes.
\item We propose a GCD classifier learning method by adaptively selecting samples via two complementary reliable sample selection methods, based on confidence scores and feature densities respectively.
\item We introduce an efficient new class number estimation method in the regime of GCD under long-tailed distribution, leveraging feature density peaks in the unlabelled data.
\end{itemize}

\section{Related Works}
\label{sec:related}

Our work is related to the fields of generalized category discovery, semi-supervised learning, and long-tailed distribution recognition, we briefly review the related works below.

\textit{Generalized Category Discovery} (GCD) is a recently formalized problem setting that aims to discover new categories in the unlabelled dataset by transferring the knowledge learned on a labelled dataset, where the unlabelled dataset may contain categories which are not present in the labelled dataset, hence the discovery of novel categories.
Early works tackle a simplified setting named Novel Category Discovery (NCD) where the assumption is that the category sets of the unlabelled dataset and the labelled dataset are dis-joint~\cite{Han2019learning,han21autonovel,hsu2017learning,hsu2019multi}.
RankStat~\cite{han21autonovel} shows that the NCD task benefits from self-supervised pretrained objectives and proposed a pair-wise objective to transfer the knowledge from the labelled set to the unlabelled dataset.
DualRank~\cite{zhao21novel} extends on this method to use local fine-grained image features and achieves a better performance on fine-grained datasets.
Contrastive Learning~\cite{zhong2021neighborhood,jia2021joint} and data augmentations~\cite{zhong2021openmix} have also been explored by previous works.
UNO~\cite{fini2021unified} proposed a unified objective to optimize for the NCD task and obtained state-of-the-art performance.
\cite{vaze22generalized} is the first work that formally introduces the GCD task, combining contrastive learning and semi-supervised $k$-means, ~\cite{vaze22generalized} can learn an effective representation on the unlabelled dataset as well as estimating an accurate class number for the whole unlabelled dataset.
Concurrent work ORCA~\cite{cao21orca} also tackles a similar setting as GCD, termed open-world semi-supervised learning.
XCon~\cite{fei2022xcon} improves upon \cite{vaze22generalized} by introducing the technique of split the training dataset into $k$-subgroups.
Recent work~\cite{wen2022simgcd} proposed a simple baseline for the GCD task and can achieve an impressive result over prior works using a parametric classifier.
Despite the progress, prior works all assume the distribution of categories in the labelled and unlabelled dataset is uniform, yet the real world may exhibit a long-tailed distribution, we are the first work that tackle the GCD task under the more realistic long-tailed setting.

\textit{Semi-supervised learning} (SSL) is a long-standing research topic that has many effective methods proposed~\cite{rebuffi2020semi,sohn2020fixmatch,berthelot2019mixmatch,laine2016temporal,tarvainen2017mean}.
The main assumption of SSL is that the unlabelled dataset shares the same set of categories with the labelled dataset, and the goal is to learn a classification model that is able to leverage the unlabelled dataset to improve its classification performance.
Self-supervised representations that can help learn a strong representation, are also shown to be effective for SSL~\cite{rebuffi2020semi,zhai2019s4l}.
Consistency methods are among the most effective methods for SSL, such as Mean-Teacher~\cite{tarvainen2017mean}, MixMatch~\cite{berthelot2019mixmatch}, and FixMatch~\cite{sohn2020fixmatch}.
Recent works shift the attention to a more realistic scenario where the assumption is that the unlabelled dataset can contain categories that are not in the labelled datasets~\cite{saito2021openmatch,yu2020multi,huang2021trash}, this setting is termed as open-set SSL. 
The main difference between open-set SSL and the GCD setting tackled in this paper is that open-set SSL simply rejects the novel categories in the unlabelled dataset without evaluating the clustering performance of those categories, while GCD is considering both seen and novel categories together to measure the performance.

\textit{Long-tailed distribution} is a long-standing problem which aims at tackling the naturally occurring long-tailed distribution in real-world datasets, where a few classes contain numerous examples (head classes) but other classes only have a few instances (tail classes). The major technical challenge in this setting is that the trained model is easily biased towards head classes and performs poorly on the tail classes. Existing works in long-tailed distribution often assume a fully-supervised setting, several techniques have been proposed, such as re-sampling~\cite{chawla2002smote,Guo_2021_CVPR,kang2019decoupling,hong2021disentangling}, re-weighting~\cite{cao2019learning,deng2021pml,he2022relieving}, logits adjustment~\cite{menon2020long,peng2022optimal,tian2021posterior} and ensembles~\cite{zhou2020BBN,wang2020long}. 
Few works focus on the long-tailed semi-supervised learning scenario, and it has been shown that similar techniques like re-sampling or re-weighting~\cite{he2021rethinking,wei2021crest,oh2021distribution,lai2022smoothed,guo2022class} still work under the semi-supervised setting.
However, these long-tailed SSL works still follow the assumption in common SSL scenarios where the unlabelled dataset contains the same set of categories as the labelled set, i.e. no novel categories in the unlabelled dataset.
In this work, we consider the case where not only the distribution of the dataset is long-tailed, but also there may exists novel categories in the unlabelled dataset.

\section{Method}
\label{sec:method}

\subsection{Preliminaries}
\subsubsection{Problem Setting}

Generalized Category Discovery (GCD) aims to learn a model for categorizing unlabelled samples in dataset $\mathcal{D}^u=\left\{(\boldsymbol{x}_i^u, \boldsymbol{y}_i^u)\right\}\in \mathcal{X}\times \mathcal{Y}_u$, using the knowledge obtained from labelled dataset $\mathcal{D}^l=\left\{(\boldsymbol{x}_i^l, \boldsymbol{y}_i^l)\right\}\in \mathcal{X}\times \mathcal{Y}_l$. 
$\mathcal{D}^u$ consists of unlabelled examples in label space $\mathcal{Y}_u$, while $\mathcal{D}^l$ contains labelled examples in label space $\mathcal{Y}_l$, where $\mathcal{Y}_l\subset\mathcal{Y}_u$. 
The number of categories in $\mathcal{Y}_u$ is denoted by $K_u$, which is typically assumed to be known a priori or can be estimated using previous methods~\cite{vaze22generalized,Han2019learning}. 
Unlike Semi-Supervised Learning (SSL) and Novel Category Discovery (NCD) settings, where $\mathcal{Y}_l=\mathcal{Y}_u$ and $\mathcal{Y}_l\cap\mathcal{Y}_u=\emptyset$, respectively, GCD is a more realistic and practical problem.
In our paper, we take a step further and assume a more realistic setting where the class distribution in the unlabelled set exhibit a long-tailed distribution.
Formally, we denote the number of examples in class $k$ as $N_k$, thus $\sum_{k=1}^{K_u}N_k=N$ where $N$ is the number of all examples.
Without the loss of generality, the classes are sorted by $N_k$ in descending order ($N_1\ge N_2\ge\dots \ge N_k$), and we can therefore represent the imbalance ratio as $\lambda=\frac{N_1}{N_k}$.

Current approaches for addressing the generalized category discovery problem typically involve two main components: representation learning and label assignment. The label assignment methods can be further subdivided into two distinct categories - parametric classifiers and non-parametric clustering methods.

In the upcoming sections, we will first present a parametric classification baseline (\cref{sec:baseline}). After that, we will introduce our proposed  methods to handle long-tailed distribution, which includes an sample selection process that optimizes classifier training to achieve balance (\cref{sec:classifier_lt}). Additionally, we will discuss a density-based class number estimation module capable of estimating the class number under the long-tailed distribution (\cref{sec:estimate_k_lt}).

\subsubsection{Baseline}
\label{sec:baseline}

We first present a strong GCD baseline proposed in~\cite{wen2022simgcd} which contains two parts, representation learning and classifier learning.

\paragraph{Representation Learning} aims to learn a general representation of all classes that can be further utilized by the classifier to classify both labelled and unlabelled classes. 
The representation learning utilizes supervised contrastive learning~\cite{supcon} $\mathcal{L}_{\text{SupCon}}$ for labelled data and self-supervised contrastive learning~\cite{chen2020simple} $\mathcal{L}_{\text{SelfCon}}$ for all the data.
The overall representation learning loss is balanced with $\lambda_{\text{rep}}$:
\begin{equation}
 \mathcal{L}_\text{rep} = (1 - \lambda_{\text{rep}}) \mathcal{L}_{\text{SelfCon}} + \lambda_{\text{rep}} \mathcal{L}_{\text{SupCon}} \,,
\end{equation}
Leveraging the powerful representation learned using contrastive methods, we can further learn a classifier for the GCD problem.

\paragraph{Classifier Learning} aims to learn a classifier for all the classes in the dataset based on the learned representations.
We can define a set of prototypes $\mathcal{C}=\{\bm{c}_1,\dots,\bm{c}_K\}$ where $K_u$ is the total number of classes in the dataset.
During training, we first calculate the predicted logits of one augmented view $\hat{\bm{x}}_i$ of the input $\bm{x}_i$ belonging to each class $k$ using the hidden features $\hat{\bm{h}}_i=f(\hat{\bm{x}}_i)$ with normlization:%
\begin{equation}
    l^{(k)}(\hat{\bm{x}}_i)=\nicefrac {(\hat{\bm{h}}_i/\|\hat{\bm{h}}_i\|_2)^\top (\bm{c}_k/ \|\bm{c}_k\|_2)}{\tau_s}.
\end{equation}
Then we use softmax to convert these logits to a probability:
\begin{equation}
    \hat{\bm{p}}_i^{(k)}=p^{(k)}(\hat{\bm{x}}_i)=\frac{\exp l^{(k)}}{\sum_{k'} \exp l^{(k')}}.
\end{equation}
We can then use the other view $\tilde{\bm{x}}_i$ of the same input $\bm{x}_i$ to calculate the soft pseudo-label $\tilde{\bm{p}}_i$ with a sharpen temperature $\tau_t$.
Then we adopted the losses proposed in~\cite{wen2022simgcd} to train the classifier:
\begin{equation}
    \mathcal{L}_{\text{cls}}^u = \frac{1}{|B|}\sum_{i\in B} \ell(\tilde{\bm{p}}_i, \hat{\bm{p}}_i) - \epsilon H(\overline{\bm{p}}),
    \mathcal{L}_{\text{cls}}^s = \frac{1}{|B^l|}\sum_{i\in B^l}\ell(\bm{y}_i, \hat{\bm{p}}_i),
\end{equation}
where $\ell(\bm{q}, \bm{p})=-\sum_k \bm{q}^{(k)}\log \bm{p}^{(k)}$ is the cross-entropy loss, and $\bm{y}_i$ is the label of $\bm{x}_i$. 
An entropy regularization is also adopted~\cite{msn}. First, we calculate the mean prediction of the a batch:
\begin{equation}
    \overline{\bm{p}} = \frac{1}{2|B|}\sum_{i \in B}\left( \hat{\bm{p}}_i+\tilde{\bm{p}}_i \right),
\end{equation} 
and the entropy $H(\overline{\bm{p}}) = -\sum_k\overline{\bm{p}}^{(k)}\log\overline{\bm{p}}^{(k)}$.
The classification loss is defined as $\mathcal{L}_{\text{cls}}=(1-\lambda_{\text{cls}})\mathcal{L}_{\text{cls}}^{u} + \lambda_{\text{cls}} \mathcal{L}_{\text{cls}}^{s}$, and the overall loss is the combination of representation learning and classifier learning losses: $\mathcal{L}_{\text{base}}=\mathcal{L}_{\text{rep}}+ \mathcal{L}_{\text{cls}}$.

\subsection{GCD Classifier for Long-tailed Distribution}
\label{sec:classifier_lt}

One major challenge that arises from the long-tailed data distribution is that the classifier may be biased towards the head classes, which have much more data than the tail classes. This could result in unreliable pseudo-labels for training and, thus, hurt the learned representation and generalization.
Our idea for tackling this challenge of training a model with a long-tailed data distribution is to leverage a sample selection method to select a balanced subset with reliable samples from the unlabelled dataset. We will use only this subset of examples to form training mini-batches and enforce the prediction distribution of the model to be close to the distribution of the selected subset.
The intuition behind this idea is that we can select a subset of high-quality data that has a roughly balanced distribution. Thus, we can help the model to migrate the bias from the original long-tailed distribution.

\begin{figure}[t]
\centering
\includegraphics[width=\columnwidth]{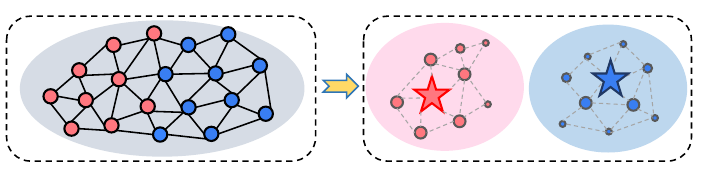}
\vspace{-15pt}
\caption{
The process of density selection. (Left) we compute the similarity between the nearest neighbors of each data sample, denoted as the edges in the figure. (Right) By using the definition in~\cref{eq:peaks}, we select a few density peaks from the raw data.
}
\label{fig:density}
\vspace{-15pt}
\end{figure}

We introduce two complementary methods for selecting reliable samples in the unlabelled dataset, one is based on the prediction confidence of the input example $\bm{x}_i$ (relying on only the individual sample), and the other one is based on using the density of each data samples~\cite{hilander} (relying on the neighbors of a sample).
Formally, for the confidence-based selection, we use the prototype classifier $p$ introduced in~\cref{sec:baseline} with the sharpened temperature $\tau_t$ to obtain the prediction of the model for each sample in the unlabelled dataset $\bm{p}_i=p(\bm{x}_i), \bm{x}_i \in \mathcal{D}^u$.
With this prediction, we sample a subset of the unlabelled data example $\mathcal{S}_{\text{conf}}$ using:
\begin{equation}
    \mathcal{S}_{\text{conf}} = \{\bm{x}_i | p(\bm{x}_i)\geq \epsilon_{\text{conf}}, \bm{x}_i \in \mathcal{D}^u \},
\end{equation}
where $\epsilon_{\text{conf}}$ denotes the threshold for the confidence selection.
For the density-based sample selection, we adopted the density definition from~\cite{hilander} which defines the density $d_i$ of a sample $\bm{x}_i$ as:
\begin{equation}
    d_i = \frac{1}{|\mathcal{N}_{\bm{x}_i}^k|} \sum_{j\in \mathcal{N}_{\bm{x}_i}^k} e_{ij} \cdot a_{ij},
\end{equation}
where $\mathcal{N}_{\bm{x}_i}^k$ is the set of $k$ nearest neighbor of the sample $\bm{x}_i$ and calculate the connectivity $e_{ij}$ of the sample $\bm{x}_i$ to its $j$-th neighbor as $e_{ij}=2 \bm{p}_i \cdot \bm{p}_j - 1$, and the affinity $a_{ij}$ as $a_{ij}=<\bm{h}_i, \bm{h}_j>$. 
Note that the choice of the density definition is not unique. Other density estimation methods are also applicable.
Intuitively, this density $d_i$ measures how compact the embedding space is  around a sample $\bm{x}_i$, the higher the density, the closer $\bm{x}_i$ to the class center.
We can consider this as a property that is irrelevant to the quality of the learned classifier, as the comparison is only done in the embedding space, thus the selected samples will not heavily influenced by the long-tailed distribution.
With the density estimation for each sample, we propose to select a set of density peaks  $\mathcal{S}_{\text{dens}}$ from the unlabelled dataset by:
\begin{equation}\label{eq:peaks}
    \mathcal{S}_{\text{dens}} = \texttt{NMDS}(\{\bm{x}_i | \forall j\in \mathcal{N}_{\bm{x}_i}^k, d_i \geq d_j, \bm{x}_i \in \mathcal{D}^u\}).
\end{equation}
Here, we first identify a set of density peaks which consist of samples with higher density than their $k$ nearest neighbors. We then use the $\texttt{NMDS}$ function for Non-Maximum Density Suppression, defined in~\cref{alg:NMDS}, to suppress redundant high-density samples in the head class. In this algorithm, the intersection-over-union function is defined as $\texttt{IoUK}(\bm{x}_i, \bm{x}_j)=\frac{\|\mathcal{N}_{\bm{x}_i}^{k_s} \cap \mathcal{N}_{\bm{x}_j}^{k_s}\|}{\|\mathcal{N}_{\bm{x}_i}^{k_s} \cup \mathcal{N}_{\bm{x}_j}^{k_s}\|}$, where $k_s$ is a hyper-parameter that sets the number of neighbors to compare in the function.
Ultimately, this process allows us to obtain $\mathcal{S}_{\text{dens}}$, a subset of the unlabelled dataset consisting of density peaks.
We provide a visualization in~\cref{fig:density} to explain the density selection process.

Together with the confidence selection, the final selected data samples form $\mathcal{S}=\mathcal{S}_{\text{conf}}\cup \mathcal{S}_{\text{dens}}$. The prior distribution $\boldsymbol{p}_{\text{prior}}$ is formed using the pseudo-label distribution within this selected subset, as reliable samples can provide a more balanced distribution for the model to learn, specifically:
\begin{equation}
    \boldsymbol{p}_{\text{prior}}=\sigma (\sum_{\boldsymbol{x} \in \mathcal{S}} \hat{y}(\boldsymbol{x})),
\end{equation}
where $\sigma$ is the softmax function, and $\hat{y}$ is a function that generate a one-hot pseudo label of the input $\boldsymbol{x}$ using the prediction $\boldsymbol{p}$.
An additional loss is then added to the model:
\begin{equation}
\mathcal{L}_{\text{prior}}=\ell(\boldsymbol{p}_{\text{prior}}, \overline{\boldsymbol{p}}),
\end{equation}
Here, $\ell$ is the cross-entropy function and $\overline{\boldsymbol{p}}$ represents the target distribution. This regularizer will drive the model to match its predicted distribution with the selected reliable sample distribution, thus improving the classifier and the underlying representations. The overall loss of the model is $\mathcal{L} = \mathcal{L}_{\text{base}} + \mathcal{L}_{\text{prior}}$.

\begin{figure*}[t]
\centering
    \includegraphics[width=\linewidth,height=0.5\columnwidth]{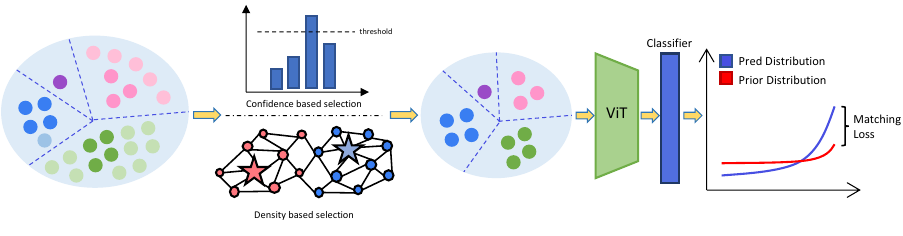}
\vspace{-25pt}
\caption{
The overall framework of classifier training.
From the original long-tailed dataset, we leverage two complementary methods to select a reliable subset of the data to form training mini-batches.
The model is trained with the subsampled data, as well as being regularized to produce a prediction distribution similar to the selected dataset.
}
\label{fig:framework}
\vspace{-10pt}
\end{figure*}

Additionally, we apply this sample selection method to the whole unlabelled dataset $\mathcal{D}^u$, selecting a subset $\hat{\mathcal{D}}^u=\mathcal{S}$ at the end of each epoch based on the same criteria used for the labelled dataset. At the start of the next epoch, we draw unlabelled training mini-batches $B$ only from the subset $\hat{\mathcal{D}}^u$. We illustrate the overall framework of our proposed method for training the classifier in~\cref{fig:framework}.

\begin{algorithm}[t]
\caption{Non-Maximum Density Suppression}\label{alg:NMDS}
\begin{algorithmic}
\Procedure {\texttt{NMDS}}{$\mathcal{S}$}
\State $\mathcal{S}_{\texttt{NMDS}} \gets \emptyset$ 
\For {$(\boldsymbol{x}_i, d_i)\in \mathcal{S}$}
    \State $\texttt{discard} \gets False$
    \For {$(\boldsymbol{x}_j, d_j)\in \mathcal{S}$}
        \If {$\texttt{IoUK}(\boldsymbol{x}_i,\boldsymbol{x}_j) > \lambda_{\texttt{NMDS}}$}
            \If {$d_i>d_j$}
                \State $\texttt{discard} \gets True$
            \EndIf
        \EndIf
    \EndFor
    \If {not \texttt{discard}}
        \State $\mathcal{S}_{\texttt{NMDS}} \gets \mathcal{S}_{\texttt{NMDS}}\cup (\boldsymbol{x}_i, d_i)$
    \EndIf
\EndFor
\EndProcedure
\Return {$\mathcal{S}_{\texttt{NMDS}}$}
\end{algorithmic}
\end{algorithm}

\subsection{Class Number Estimation}
\label{sec:estimate_k_lt}

Another major challenge with the long-tailed distribution in generalized category discovery is that it can be hard to estimate the number of classes in the unlabelled set using the conventional semi-supervised $k$-means~\cite{vaze22generalized} algorithm, as the $k$-means algorithm assumes that each of the clusters in the data is isotropic with roughly equal number of samples per cluster~\cite{liang2012k,wu2009adapting}. 
This assumption, however, does not hold in the long-tailed distribution we want to tackle, thus we cannot directly adopt the semi-supervised $k$-means algorithm for estimating the number of classes in the long-tailed GCD setting.
Here we propose a novel algorithm for determining the number of categories in the unlabelled set using the concept of the $k$-NN density of data samples~\cite{hilander}.
The main idea is similar to the density-based selection, higher density and density peak samples are more likely to be closer to the cluster center of a cluster~\cite{ester1996density,hilander}, thus we can measure the potential number of classes in the unlabelled data by counting how many density peak samples are there in the data.

Formally, given a dataset of samples $\mathcal{D}^u=\{(\bm{x}_i^u, \bm{y}_i^u)\}$, our goal is to determine the number of classes $K_u$ in $\mathcal{D}^u$ as well as an assignment $\hat{\bm{y}_i}$ of each data samples $\bm{x}_i^u$.
Instead of iteratively calculating the class prototypes and updating class assignments, we take the concept of a `density peak' in the dataset~\cite{ester1996density,hilander} to implement an algorithm for estimating $K_u$.
A `density peak' $\bm{x}_i$ is defined as the density $d_i$ is higher than all the $k$ neighbours of $\bm{x}_i$, $\forall j \in \mathcal{N}_{\bm{x}_i}, d_i > d_j$.
After getting all the density peaks in the dataset $\mathcal{D}^u$, we run an $\texttt{NMDS}$ step using the algorithm defined in~\cref{alg:NMDS} to remove duplicated density peaks that may belong to the same categories.
The class assignment is done by using those density peaks as class prototypes and performing a simple distance-based assignment.
We can then obtain an upper bound of the number of classes as the total number of density peaks and the lower bound as the number of labelled categories. With these two bounds, we adopt Brent's algorithm~\cite{brent} to find the optimal new category number by clustering with our density-based approach above on the mixed set of labelled and unlabelled data. During the process, we drop the labels of the labelled data. The optimal value is obtained by examining the optimal clustering accuracy on the labelled data.

\section{Experiments}
\label{sec:exp}

\subsection{Experimental Setup}
\label{sec:exp-setup}

\paragraph{Benchmark and Evaluation Metrics.}
We validate the performance of our methods on long-tailed datasets, including long-tailed CIFAR-100~\cite{cifar}, ImageNet-100~\cite{deng2009imagenet} as well as naturally occuring long-tailed datasets including Herbarium-19~\cite{tan2019herbarium} and iNat-18~\cite{inat}.
For a comprehensive evaluation of the proposed methods, we also conduct experiments on uniformed distributed datasets for GCD, including ImageNet-100~\cite{deng2009imagenet} and the Semantic Shift Benchmark(SSB)~\cite{vaze22openset} which includes CUB~\cite{cub200} and Stanford Cars~\cite{stanfordcars}.
For each of the datasets, we follow previous works~\cite{vaze22generalized,wen2022simgcd} to sample a subset of all classes as the old classses $\mathcal{Y}_l$; 
50\% of the images from these labelled classes are used to construct $\mathcal{D}^l$, and the remaining images are regarded as the unlabelled data $\mathcal{D}^u$. 
See~\cref{tab:datasplit} for statistics of the datasets we evaluate on, as well as the imbalance factor $\lambda$ of each datasets.
The model is evaluated using the clustering accuracy (ACC) following the standard practice~\cite{vaze22openset,wen2022simgcd}.
For the long-tailed dataset, we compute the balanced-ACC as the average of per-class ACC for an unbiased evaluation.

\begin{table*}[ht]
\centering
\caption{    
Results on long-tailed distribution datasets. \label{tab:lt}
}
\resizebox{.9\linewidth}{!}{ %
\tablestylesmaller{1pt}{1.05}
\begin{tabular}{lllllllllll}
\toprule
 & &    \multicolumn{3}{c}{CIFAR100-LT} & \multicolumn{3}{c}{ImageNet100-LT} & \multicolumn{3}{c}{Herb-19} \\
\cmidrule(r){3-5}\cmidrule(r){6-8}\cmidrule(r){9-11}
No. & Methods                                   & \,All  & \,Old  & New  & \,All  & \,Old  & New  & \,All  & \,Old  & New \\\midrule
(1) & $k$-means~\cite{macqueen1967some_kmeans}  & 31.3 & 35.3 & 30.1 & 51.9 & 67.2 & 30.8     & 13.6 & 12.2 & 15.0 \\
(2) & RankStats+~\cite{han21autonovel}          & 45.7 & 59.1 & 24.1 & 47.4 & 70.1 & 23.1     & 11.4 & 13.2 & 12.5 \\
(3) & UNO+~\cite{fini2021unified}               & 49.9 & 61.1 & 25.7 & 51.2 & 74.2 & 25.1     & 15.3 & 17.1 & 13.4 \\
(4) & ORCA~\cite{cao21orca}                     & 41.3 & 59.5 & 20.7 & 46.3 & 67.1 & 24.2     & \,\,9.8 & 14.7 & \,\,4.9 \\
(5) & GCD~\cite{vaze22generalized}              & 62.3 & 66.9 & 28.1 & 53.1 & 75.1 & 28.3     & 32.8 & 41.4 & 24.2 \\
(6) & SimGCD~\cite{wen2022simgcd}               & \uline{70.4} & \uline{77.4} & \uline{32.1} & \uline{56.6} & \uline{79.6} & \uline{33.5}     & \uline{39.4} & \uline{51.4} & \uline{27.3}      \\
\midrule
(7) & Ours                     & \textbf{72.1}\tiny{$\pm$0.4} & \textbf{80.1}\tiny{$\pm$1.2} & \textbf{33.7}\tiny{$\pm$0.8} & \textbf{58.4}\tiny{$\pm$0.2} & \textbf{83.1}\tiny{$\pm$0.5} & \textbf{35.4}\tiny{$\pm$1.5} & \textbf{43.5}\tiny{$\pm$0.9} & \textbf{55.8}\tiny{$\pm$0.6} & \textbf{28.5}\tiny{$\pm$0.4} \\
\bottomrule
\end{tabular}
}
\end{table*}

\begin{table}[htbp]   
    \centering
    \tablestyle{4pt}{1}
    \caption{Statistics of the datasets we evaluate on.}\label{tab:datasplit}
    \begin{tabular}{lrcrcc}
    \toprule
                & \multicolumn{2}{c}{Labelled}  & \multicolumn{2}{c}{Unlabelled} & \\
                \cmidrule(rl){2-3}\cmidrule(rl){4-5}
    Dataset            & \#Image   & \#Class   & \#Image   & \#Class  & $\lambda$\\
    \midrule
    CUB~\cite{cub200}                 & 1.5K      & 100       & 4.5K      & 200  & 1.0 \\
    Stanford Cars~\cite{stanfordcars}       & 2.0K      & 98        & 6.1K      & 196  & 1.0\\    
    ImageNet-100~\cite{deng2009imagenet}       & 31.9K     & 50        & 95.3K     & 100  & 1.0\\
    \midrule
    CIFAR-100-LT~\cite{cifar}        & 4K      & 80       & 11K     & 100 & 10.0\\
    ImageNet-100-LT~\cite{deng2009imagenet}       & 6K     & 50        & 22K     & 100 & 10.0\\
    \midrule
    Herbarium 19~\cite{tan2019herbarium}        & 9K      & 341       & 25K     & 683  & 46.1 \\
    iNatualist-18~\cite{inat}        & 130K      & 4,071        & 307K      & 8,142 & 500.0\\
    \bottomrule
    \end{tabular}
    \vspace{-1em}
\end{table}

\paragraph{Implementation Details.}
Following the common practice, we train all methods with the ViT-base/16 model~\cite{dosovitskiy2020vit} pretrained with DINO~\cite{caron2021emerging}.
The \texttt{[CLS]} token with a dimension of 768 is used as the feature representation of one image and we only finetuned the last block of the backbone. 
The model is trained with a batch-size of 128, with an initial learning rate of 0.1 decayed with a cosine schedule to 0.
And $\epsilon_{\text{conf}}$ is set to $0.8$.
For a fair comparison, we train for 200 epochs on each dataset, and the best-performing model is selected using the accuracy of the validation set of the labelled classes.

\subsection{Comparison with the State-of-the-Art}

We present a comparison of our method with state-of-the-art methods on both long-tailed datasets(~\cref{tab:lt}) and the SSB benchmark datasets(~\cref{tab:ssb}).
From~\cref{tab:lt} we can observe that our method achieves the overall best performance on the challenging long-tailed distribution datasets, outperforming the second-best model SimGCD~\cite{wen2022simgcd} by 1.1\%-4.4\% in ACC, validating the effectiveness of our method for handling the long-tailed distribution.
It can also be observed that our method demonstrates a non-trivial performance improvement over them compared with other previous state-of-the-art.
Comparing the $k$-means baseline with other methods, we can see that only GCD~\cite{vaze22generalized}, SimGCD~\cite{wen2022simgcd}, and our method consistently outperforms the baseline on the performance on the long-tailed `New' categories, this demonstrate the difficulty when dealing with the long-tailed distribution for categories discovery. In~\cref{tab:lt}, our method achieves the best performance in all cases. 
In~\cref{tab:ssb}, our method achieves the best performance on `all' and `new' classes, while performing on par with SimGCD on `old' classes. 
Note that our method is designed to handle the challenging long-tailed scenario, thus achieving a performance that is on par with the performance of strong methods on conventional benchmarks is an encouraging result that indicates that our method can be used for different scenarios.

\begin{table*}[t]
\centering
\caption{
Results on the Semantic Shift Benchmark~\cite{vaze22openset}. \label{tab:ssb}
}
\resizebox{.9\linewidth}{!}{ %
\tablestylesmaller{1pt}{1.05}
\begin{tabular}{lllllllllll}
\toprule
&&   \multicolumn{3}{c}{CUB} & \multicolumn{3}{c}{Stanford Cars} & \multicolumn{3}{c}{ImageNet-100}\\
\cmidrule(r){3-5}\cmidrule(r){6-8}\cmidrule(r){9-11}
No. & Methods        & \,All  & \,Old  & New  & \,All  & \,Old  & New  & \,All  & \,Old  & New \\\midrule
(1) & $k$-means~\cite{macqueen1967some_kmeans}  & 34.3 & 38.9 & 32.1 & 12.8 & 10.6 & 13.8  & 72.7 & 75.5 & 71.3 \\
(2) & RankStats+~\cite{han21autonovel}          & 33.3 & 51.6 & 24.2 & 28.3 & 61.8 & 12.1 & 37.1 & 61.6 & 24.8 \\
(3) & UNO+~\cite{fini2021unified}               & 35.1 & 49.0 & 28.1 & 35.5 & 70.5 & 18.6 & 70.3 & \textbf{95.0} & 57.9 \\
(4) & ORCA~\cite{cao21orca}                     & 35.3 & 45.6 & 30.2 & 23.5 & 50.1 & 10.7 & 73.5 & \uline{92.6} & 63.9 \\
(5) & GCD~\cite{vaze22generalized}              & 51.3 & 56.6 & 48.7 & 39.0 & 57.6 & 29.9 & 74.1 & 89.8 & 66.3 \\
(6) & SimGCD~\cite{wen2022simgcd}               & \uline{60.3} & \textbf{65.6} & \uline{57.7} & \uline{46.8} & \textbf{64.9} & \uline{38.0}  & \textbf{82.4} & 90.7 & \textbf{78.3}  \\
\midrule
(7) & Ours                     & \textbf{61.3}\tiny{$\pm$0.1} & \uline{64.2}\tiny{$\pm$0.9} & \textbf{59.2}\tiny{$\pm$0.4} & \textbf{47.9}\tiny{$\pm$1.8} & \uline{64.7}\tiny{$\pm$1.3} & \textbf{39.3}\tiny{$\pm$2.1} & \uline{81.1}\tiny{$\pm$2.2} & 88.4\tiny{$\pm$2.2} & \uline{77.8}\tiny{$\pm$2.7} \\
\bottomrule
\end{tabular}
}
\end{table*}

\subsection{Novel Class Number Estimation}

In~\cref{tab:k_est}, we show the performance of estimating the number of categories in the unlabeled dataset.
We first show a comparison of estimated category numbers on uniformed datasets including CIFAR-100 and ImageNet-100, compared with the search algorithm proposed in~\cite{vaze22generalized}, our method gives comparable estimation performance.
Importantly, for the real-world long-tailed distribution datasets including artificial splitted long-tailed datasets like CIFAR-100-LT and ImageNet-100-LT and naturally occurring long-tailed datasets like Herb-19 and iNat-18, the estimation of our method exhibits non-trivial improvements over the method in~\cite{vaze22generalized}.
These results indicate the effectiveness of our method when applied to the long-tailed datasets.

\begin{table}[t]
\centering
\caption{
Estimation of class numbers in unlabelled data.
}
\label{tab:k_est}
\resizebox{0.8\linewidth}{!}{ %
\begin{tabular}{l|ccc}
\toprule
Dataset & GT & GCD~\cite{vaze22generalized} & Ours \\
\midrule
CIFAR100        & 100 & \textbf{100} & 109 \\
ImageNet-100    & 100 & \textbf{109} & 112 \\
\midrule
CIFAR100-LT     & 100 & 78 & \textbf{86} \\
ImageNet-100-LT & 100 & 71 & \textbf{79} \\
Herb-19         & 683 & 520 & \textbf{586} \\
iNat-18         & 8,142 & 5981 & \textbf{6,151} \\
\bottomrule
\end{tabular}
} %

\end{table}

\subsection{Ablation Study}

In this section, we provide ablations to each component of our method.

\paragraph{Performance with different imbalance factors $\lambda$.}
Firstly, we present an ablation to study the performance variation when the imbalance factor $\lambda$ is different.
We use CIFAR-100 and ImageNet-100 to create artificial splits with different imbalance factors by subsampling the original dataset.
The clustering results are presented in~\cref{tab:lambda}, and these demonstrate that our method consistently produces better outcomes across the different $\lambda$ values that we tested. Moreover, we observed that when $\lambda$ increases, the performance gap between our proposed method and SimGCD also expands. This result highlights the effectiveness of our proposed method in handling the long-tailed distribution.

The results of estimating the category numbers are presented in~\cref{tab:lambda_k_est}, which reveal that as $\lambda$ increases, the estimated number decreases due to the fact that the smallest cluster becomes smaller with higher imbalance, and it is more likely for the estimation algorithm to overlook it. According to the results, our proposed method outperforms the algorithm in~\cite{vaze22generalized} in all scenarios.

\begin{table}[t]
\centering
\caption{Results of evaluating using different imbalance factors $\lambda$. All results are in `All / Old / New' format.}
\label{tab:lambda}
\begin{tabular}{ccc}
\toprule
CIFAR-100-LT & SimGCD & Ours \\
\midrule
$\lambda=5$  & 73.5 / 80.2 / 35.1 & 74.2 / 81.0 / 35.6 \\
$\lambda=10$ & 70.4 / 77.4 / 32.1 & 72.1 / 80.1 / 33.7 \\
$\lambda=20$ & 63.1 / 70.3 / 28.6 & 67.2 / 75.3 / 30.2 \\
\midrule
ImageNet-100-LT & SimGCD & Ours \\
\midrule
$\lambda=5$  & 62.1 / 83.1 / 37.1 & 63.1 / 84.5 / 38.1 \\
$\lambda=10$ & 56.6 / 79.6 / 33.5 & 58.4 / 83.1 / 35.4 \\
$\lambda=20$ & 50.1 / 74.5 / 26.3 & 54.2 / 78.2 / 29.1 \\
\bottomrule
\end{tabular}
\end{table}

\begin{table}[t]
\centering
\caption{Results of estimated class number for different $\lambda$.}
\label{tab:lambda_k_est}
\begin{tabular}{ccc}
\toprule
CIFAR-100-LT & GCD & Ours \\
\midrule
$\lambda=5$  & 87 & \textbf{89} \\
$\lambda=10$ & 78 & \textbf{86} \\
$\lambda=20$ & 65 & \textbf{80} \\
\midrule
ImageNet-100-LT & GCD & Ours \\
\midrule
$\lambda=5$  & 85 & \textbf{87} \\
$\lambda=10$ & 71 & \textbf{79} \\
$\lambda=20$ & 60 & \textbf{73} \\
\bottomrule
\end{tabular}
\end{table}

\paragraph{Confidence sample selection.}
In~\cref{tab:s_option}, we show an ablation study using different combinations of the selected subset $\mathcal{S}$.
The default choice is to use $\mathcal{S}_{\text{conf}}\cup \mathcal{S}_{\text{dens}}$ as $\mathcal{S}$ for forming training mini-batches to train the model. 
Here we explore the performance of only using $\mathcal{S}_{\text{conf}}$ or $\mathcal{S}_{\text{dens}}$ as $\mathcal{S}$.
From~\cref{tab:s_option}, we can observe that removing any one of $\mathcal{S}_{\text{conf}}$ or $\mathcal{S}_{\text{dens}}$ results in a performance degradation.
The performance degrades the most when $\mathcal{S}_{\text{conf}}$ is removed from $\mathcal{S}$, the gap is about \~10\% on `All', `Old', and `New' categories.
These results demonstrate that both $\mathcal{S}_{\text{conf}}$ and $\mathcal{S}_{\text{dens}}$ are essential to the final performance of the model validating the design choice of our method.

\begin{table}[t]
\centering
\caption{Results of varying selected samples $\mathcal{S}$.}
\label{tab:s_option}
\begin{tabular}{ccc}
\toprule
&  ImageNet-100-LT & Herb-19 \\
\midrule
Ours w/o $\mathcal{S}_{\text{conf}}$ &  40.2 / 71.2 / 22.5 & 30.0 / 38.5 / 10.4 \\
Ours w/o $\mathcal{S}_{\text{dens}}$ &  55.1 / 80.2 / 31.4 & 40.2 / 51.4 / 24.6 \\
\midrule
Ours                                 &  \textbf{58.4} / \textbf{83.1} / \textbf{35.4} & \textbf{43.5} / \textbf{55.8} / \textbf{28.5} \\
\bottomrule
\end{tabular}
\vspace{-10pt}
\end{table}

\paragraph{How balanced is the selected subset?}
We show the imbalance factor $\lambda$ of the distribution within the selected subset $\mathcal{S}$.
These statistics are shown in~\cref{tab:lambda_s}.
The original imbalance factor $\lambda$ of the whole unlabeled dataset is shown in the first row.
We can observe from the following rows that using only the confidence-based selection to select samples for $\mathcal{S}_{\text{conf}}$ is not able to reduce the imbalanced distribution of the dataset.
Using the density-based selection can indeed sample a more balanced subset from the original dataset, yet as shown in~\cref{tab:s_option}, using $\mathcal{S}_{\text{dens}}$ alone can not achieve good performance for category discovery.
Thus we need to combine these two subsets to form $\mathcal{S}=\mathcal{S}_{\text{conf}}\cup\mathcal{S}_{\text{dens}}$ to enjoy the benefit of a more balanced training set and a better performance simultaneously.
The combined $\mathcal{S}$ has a more balanced dataset than the original dataset measured by the imbalanced factor.

\begin{table}[t]
\centering
\caption{Imbalance factor $\lambda$ within the selected subset $\mathcal{S}$.}
\label{tab:lambda_s}
\begin{tabular}{ccc}
\toprule
 &  ImageNet-100-LT & Herb-19 \\
\midrule
$\mathcal{D}^u$                                                    & 10.0 & 46.1 \\
\midrule
$\mathcal{S}_{\text{conf}}$                                        & 8.4  & 44.3 \\
$\mathcal{S}_{\text{dens}}$                                        & \textbf{2.0}  & \textbf{10.2} \\
$\mathcal{S} = \mathcal{S}_{\text{conf}}\cup \mathcal{S}_{\text{dens}}$  & 4.5 & 20.7 \\
\bottomrule
\end{tabular}
\end{table}

\paragraph{Effect of the \texttt{NMDS} algorithm}
We validate the effectiveness of our proposed $\texttt{NMDS}$ algorithm by removing it and evaluating the performance in~\cref{tab:nmds}.
Comparing the first two lines in~\cref{tab:nmds}, we can see that removing the $\texttt{NMDS}$ algorithm would result in a performance drop. 
To investigate this phenomenon further, we show the imbalance factor $\lambda_{\mathcal{S}}$ of the selected subset $\mathcal{S}$ in the bottom two lines of~\cref{tab:nmds}.
We can see that without the use of the $\texttt{NMDS}$ algorithm, the imbalance factor $\lambda_{\mathcal{S}}$ would be significantly higher than when we use the $\texttt{NMDS}$ algorithm.

\begin{table}[t]
\centering
\caption{Effectiveness of the $\texttt{NMDS}$ algorithm.}
\label{tab:nmds}
\begin{tabular}{ccc}
\toprule
 &  ImageNet-100-LT & Herb-19 \\
\midrule
Ours w/o $\texttt{NMDS}$     & 55.1 / 76.2 / 31.8 & 40.1 / 50.7 / 25.1 \\
Ours w/ $\texttt{NMDS}$      & 58.4 / 83.1 / 35.4 & 43.5 / 55.8 / 28.5 \\
\midrule
$\lambda_{\mathcal{S}}$ w/o $\texttt{NMDS}$     & 7.8 & 36.5 \\
$\lambda_{\mathcal{S}}$ w/ $\texttt{NMDS}$      & 4.5 & 20.7 \\
\bottomrule
\end{tabular}
\vspace{-15pt}
\end{table}

\paragraph{Number of Nearest Neighbours}
We use two numbers of nearest neighbours in our proposed method, one is the number $k$ used as the number of nearest neighbours for calculating the density and density peaks, the other one is the number of $k_s$ used in the $\texttt{IoUK}$ function to determine the overlap between two different density peaks.
We first ablate on the influence of $k$ for the clustering performance, intuitively, a larger $k$ can cover more neighbors, thus providing a more accurate estimation of density peaks. However, by covering a larger neighbourhood, we would expect the selected number of density peaks to drop as it is less like for one sample to have a higher density than a larger number of neighbors (In the extreme scenario where $k$ is the number of total samples in a dataset, there will only be one density peak). 
In~\cref{tab:k_clustering}, we can observe that the optimal value for $k$ in our experiments is around $10$, and this value is consistent across the ImageNet-100-LT and Herb-19 datasets. We set $10$ as the default choice in our method, though the sensitivity to different numbers is not high.

\begin{table}[t]
\centering
\caption{Ablation of $k$ for clustering performance.}
\label{tab:k_clustering}
\begin{tabular}{ccc}
\toprule
$k$ &  ImageNet-100-LT & Herb-19 \\
\midrule
5       & 56.1 / 82.4 / 33.8 & 41.2 / \textbf{56.1} / 27.1 \\
10      & \textbf{58.4} / \textbf{83.1} / \textbf{35.4} & \textbf{43.5} / 55.8 / 28.5 \\
15      & 57.0 / 81.9 / 33.0 & 42.1 / 54.1 / \textbf{29.4} \\
20      & 55.2 / 80.1 / 31.2 & 40.1 / 52.4 / 26.9 \\
\bottomrule
\end{tabular}
\end{table}

\begin{table}[t]
\centering
\caption{Ablation of $k_s$ for category number estimation performance.}
\label{tab:k_s_estimation}
\begin{tabular}{ccc}
\toprule
$k_s$ &  ImageNet-100-LT & Herb-19 \\
\midrule
GT       & 100 & 683 \\
\midrule
10       & 146 & 761 \\
20       & 135 & 620 \\
30       & 79 & 586 \\
40       & 65 & 511 \\
\bottomrule
\end{tabular}
\vspace{-15pt}
\end{table}

In~\cref{tab:k_s_estimation}, we show the ablation on $k_s$ evaluting on the performance for estimating the number of categories.
We can observe that when the value of $k_s$ is small, the $\texttt{IoUK}$ function can only cover a small region in the embedding to perform $\texttt{NMDS}$, thus the method tends to over-estimate the number of categories in the dataset.
When the value of $k_s$ is larger than the optimal value, the $\texttt{IoUK}$ function will overestimate the similarity between two density peaks, thus the method could make more false negative removal of density peaks, leading to an underestimate of the categories.
In our experiments, we set $k_s$ to $30$ as the default value.

\section{Conclusion}
\label{sec:conclusion}
In this paper, we addressed the challenge of generalized category discovery in a long-tailed distribution, a problem that has not been explored in previous works. We identified two key technical challenges - balancing the classifier for all categories and estimating the category numbers accurately in the presence of a long tail. To overcome these challenges, we proposed a novel method based on sample densities. Our approach iteratively selects a balanced and reliable subset from the original unlabelled dataset and estimates category numbers using density-based clustering. Our experiments on both long-tailed and uniform datasets demonstrate the effectiveness of our method in discovering novel categories accurately. Overall, our approach provides a valuable contribution to the field of long-tailed GCD and opens up new directions for future research.

{\small
\bibliographystyle{ieee_fullname}
\bibliography{egbib}
}

\appendix
\setcounter{table}{0}
\renewcommand{\thetable}{A\arabic{table}}
\setcounter{figure}{0}
\renewcommand{\thefigure}{A\arabic{figure}}

\section{Additional ablations}

Here we present additional ablation studies for our proposed method.

\subsection{Ablation on $\lambda_\texttt{NMDS}$}

We first present the ablation on the value of the threshold $\lambda_\texttt{NMDS}$. The results are shown in~\cref{tab:lambda_nmds_estimation}. 
Similar to the ablation of varying $k_s$ in Table 11 of the main paper, setting $\lambda_\texttt{NMDS}$ to a lower value will result in an overestimation of the number of categories, and a higher value will result in an underestimation of categories.

\begin{table}[ht]
\centering
\caption{Ablation of $\lambda_\texttt{NMDS}$ for category number estimation performance.}
\label{tab:lambda_nmds_estimation}
\begin{tabular}{ccc}
\toprule
$\lambda_{\texttt{NMDS}}$ &  ImageNet-100-LT & Herb-19 \\
\midrule
GT       & 100 & 683 \\
\midrule
0.2       & 167 & 743 \\
0.4       & 153 & 617 \\
0.6       & 113 & 594 \\
0.8       & 85  & 523 \\
\bottomrule
\end{tabular}
\vspace{-15pt}
\end{table}

\subsection{Ablation on different choices of density calculation}

Here, we compare different choices of the feature density formulation. In the main paper, we use both the connectivity $e_{ij}$ between two data point $\bm{x}_i$ and $\bm{x}_j$, as well as the affinity $a_{ij}$ between the two samples.
This definition of density is inherited from HiLander ~\cite{hilander}. In this section, we also experiment on an alternative definition of the feature density, which simply averages the affinity of samples.
The density is defined as:
\begin{equation}
    d_i = \frac{1}{|\mathcal{N}_{\bm{x}_i}^k|} \sum_{j\in \mathcal{N}_{\bm{x}_i}^k} a_{ij}.
\end{equation}
We present the results in~\cref{tab:density}. We can see that if we only use the affinity for calculating the density, the results will be far worse than using connectivity and affinity together.
We conjecture this to the fact that our evaluation is under the long-tailed distribution, thus using affinity alone cannot give a good estimation of the density for tail classes and this leads to the degradation in performance.

\begin{table}[t]
\centering
\caption{Comparison of different feature density formulations.}
\label{tab:density}
\begin{tabular}{ccc}
\toprule
 &  ImageNet-100-LT & Herb-19 \\
\midrule
Only Affinity & 40.7 / 64.3 / 10.8 & 23.4 / 41.5 / 9.8 \\
Ours          & \textbf{58.4} / \textbf{83.1} / \textbf{35.4} & \textbf{43.5} / \textbf{55.8} / \textbf{28.5} \\
\bottomrule
\end{tabular}
\vspace{-15pt}
\end{table}

\section{Combine with Long-tailed Semi-Supervised Learning}

Our method tackles the challenge of learning generalized category discovery under the long-tailed distribution.  Long-tailed semi-supervised learning is a neighboring problem to ours, which also assumes a long-tailed distribution of classes. However, our setting differs from long-tailed semi-supervised learning since we have novel categories within the unlabelled set that cannot be directly handled by long-tailed semi-supervised learning methods, which cannot handle novel categories.
In this section, we propose to combine our method with long-tailed semi-supervised learning to further boost the performance of our method on the task of generalized category discovery. Specifically, we adopt CReST~\cite{wei2021crest}, a popular baseline in long-tailed semi-supervised learning that adjusts the threshold for sampling different categories based on their frequency. To combine CReST with our method, we use their sampling technique to sample our $\mathcal{S}_{\text{conf}}$ set by varying the threshold for different categories.
The results are presented in~\cref{tab:crest}. As can be seen, introducing the long-tailed semi-supervised learning techniques into our method indeed leads to better performance.

\begin{table}[t]
\centering
\caption{Combine with long-tailed semi-supervised learning method.}
\label{tab:crest}
\begin{tabular}{ccc}
\toprule
 &  ImageNet-100-LT & Herb-19 \\
\midrule
Ours          & 58.4 / 83.1 / 35.4 & 43.5 / 55.8 / 28.5 \\
\midrule
Ours+CReST    & 59.4 / 84.6 / 36.7 & 44.6 / 57.4 / 29.4 \\
\bottomrule
\end{tabular}
\end{table}

\section{Runtime for Category Number Estimation}
We compare the runtime between our method and the previous SOTA method~\cite{vaze22generalized}. The result is presented in~\cref{tab:runtime}. Our method is more than 30$\times$ faster than ~\cite{vaze22generalized}, while achieving more accurate category number estimation on the long-tailed datasets ImageNet-100-LT and Herb-19 (see Table 5 in the main paper).

\begin{table}[t]
\centering
\caption{Runtime comparison for category number estimation.}
\label{tab:runtime}
\begin{tabular}{ccc}
\toprule
 &  ImageNet-100-LT & Herb-19 \\
\midrule
Vaze~\etal~\cite{vaze22generalized} & 35,624s  & 63,901s   \\
Ours                                & 1,192s  & 1,874s   \\
\bottomrule
\end{tabular}
\end{table}

\end{document}